\documentclass{article}

\PassOptionsToPackage{numbers, compress}{natbib}

\usepackage[preprint]{neurips_2024}

\usepackage[utf8]{inputenc} 
\usepackage[T1]{fontenc}    
\usepackage{booktabs}       
\usepackage{amsfonts}       
\usepackage{nicefrac}       
\usepackage{microtype}      
\usepackage{xcolor}         
\usepackage{amsmath}
\usepackage{graphicx}
\renewcommand{\footnote}[1]{}
\usepackage{hyperref}       
\hypersetup{colorlinks,allcolors=green}

\usepackage{natbib}

\title{SA-GS: Semantic-Aware Gaussian Splatting for Large Scene Reconstruction with Geometry Constrain}

\author{%
Butian Xiong \\
CUHK, Shenzhen\\
\texttt{butianxiong@link.cuhk.edu.cn}\\
\vspace{-1em} \And
Xiaoyu Ye \\
Beijing Institute of Technology\\
\texttt{1120201575@bit.edu.cn}\\
\vspace{-1em} \And
Tze Ho Elden Tse \\
Auki Labs \\
\texttt{txt994@student.bham.ac.uk}\\
\vspace{-1em} \And
Kai Han\\
The University of Hong Kong\\
\texttt{kaihanx@hku.hk}\\
\vspace{-1em} \And
Shuguang Cui \\
CUHK, Shenzhen\\
\texttt{shuguangcui@cuhk.edu.cn}\\
\vspace{-1em} \And
Zhen Li \\
CUHK, Shenzhen\\
\texttt{lizhen@cuhk.edu.cn}\\
}

\begin{document}

\maketitle
\begin{figure*}[htbp]
  \centering
    \includegraphics[width=\linewidth]{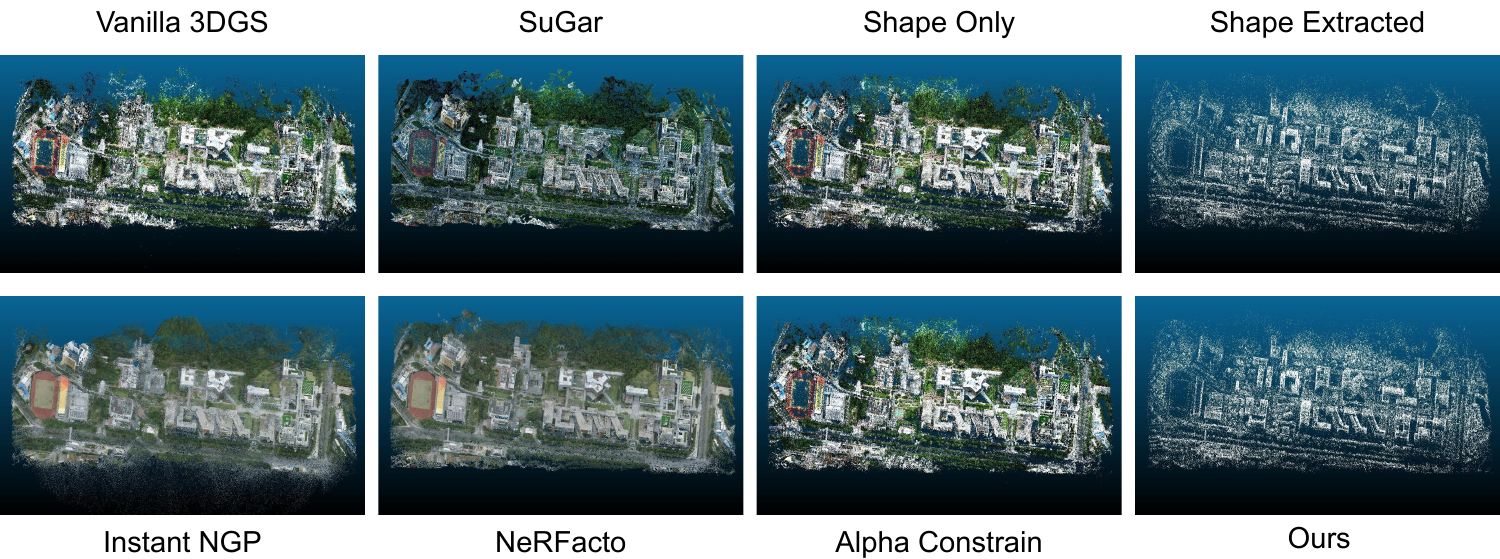}
   \caption{Qualitative comparison between our method and other 3DGS based methods. We proposed Shape constrain, alpha constrain and point cloud extraction in the current study. Quantitative ablation is shown in the right handside of the figure.}
   \label{fig:Geometric_Comparison}
\end{figure*}

\begin{abstract}
With the emergence of Gaussian Splats, recent efforts have focused on large-scale scene geometric reconstruction. However, most of these efforts either concentrate on memory reduction or spatial space division, neglecting information in the semantic space. In this paper, we propose a novel method, named SA-GS, for fine-grained 3D geometry reconstruction using semantic-aware 3D Gaussian Splats. Specifically, we leverage prior information stored in large vision models such as SAM and DINO to generate semantic masks. We then introduce a geometric complexity measurement function to serve as soft regularization, guiding the shape of each Gaussian Splat within specific semantic areas. Additionally, we present a method that estimates the expected number of Gaussian Splats in different semantic areas, effectively providing a lower bound for Gaussian Splats in these areas. Subsequently, we extract the point cloud using a novel probability density-based extraction method, transforming Gaussian Splats into a point cloud crucial for downstream tasks. Our method also offers the potential for detailed semantic inquiries while maintaining high image-based reconstruction results. We provide extensive experiments on publicly available large-scale scene reconstruction datasets with highly accurate point clouds as ground truth and our novel dataset. Our results demonstrate the superiority of our method over current state-of-the-art Gaussian Splats reconstruction methods by a significant margin in terms of geometric-based measurement metrics. Code and additional results will soon be available on our  \href{https://saliteta.github.io/SA-GS}{project page}
\end{abstract}

\section{Introduction}
\begin{figure*}
    \centering
    \includegraphics[width=\linewidth]{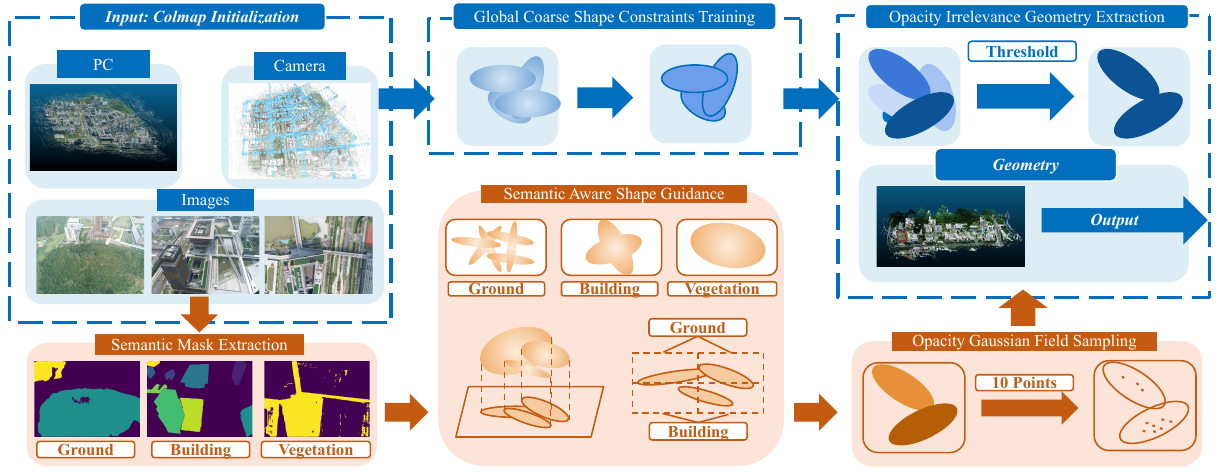}
    \caption{Overview: The blue section of the figure illustrates common methods for reconstructing geometrically aligned Gaussian Splats. The input for all Gaussian Splatting methods includes a COLMAP initialization consisting of images, camera positions, and SfM sparse point clouds. The output will be a traditional representation such as a mesh or point cloud, as shown in the right blue box. During training, in addition to the common image rendering loss, most methods encourage all 3D Gaussians to form a disk-like shape, as seen in \cite{SuGaR} and \cite{2DGS}. After several training iterations, or at the end of the training process, other methods select a hard threshold for the alpha value and use the remaining Gaussians for geometric reconstruction. However, these hard constraints often result in poorer reconstruction, as demonstrated in our experiments. Instead of encouraging all Gaussians to adopt the same shape, our method uses semantic information to control the shape in detail. We first produce semantic masks for each input image, then extract shape information for each semantic group, and use this information to locally control the shape of each Gaussian. Additionally, we provide an opacity field sampling method that can dynamically allocate the desired number of points and ignore defective reconstruction parts.}
    \label{fig:SAM_GS_Teaser}
\end{figure*}
3D reconstruction is a transformative technology that converts real-world scenes into digital three-dimensional models. 
As this technology often requires the transformation of multiple 2D images into 3D models, it finds numerous applications in urban planning, virtual reality (VR), and augmented reality (AR).
Various techniques have been employed to enhance the accuracy and efficiency of 3D reconstruction, such as Neural Field-based methods \cite{NeRFusion}\cite{coordinateNeRF}\cite{mipsnerf}\cite{NeRFrendering}\cite{vanilla_nerf}\cite{nvs_nerf} and Gaussian Splatting-based methods \cite{vanilla_gaussian}\cite{MeshInitGaussian}\cite{gaussian_pro}\cite{scaffoldGaussian}\cite{SuGaR}.


Neural Rendering techniques, such as NeRF, often face challenges due to their lengthy training times and the need for dense camera poses. These factors make them difficult to train, render and edit. In contrast, 3D Gaussian Splatting (3DGS) \cite{vanilla_gaussian} combines rasterization with novel view synthesis which features rapid training and rendering speeds and exhibits high tolerance to sparse camera positions and orientations.

\begin{figure*}
    \centering
    \includegraphics[width=\linewidth]{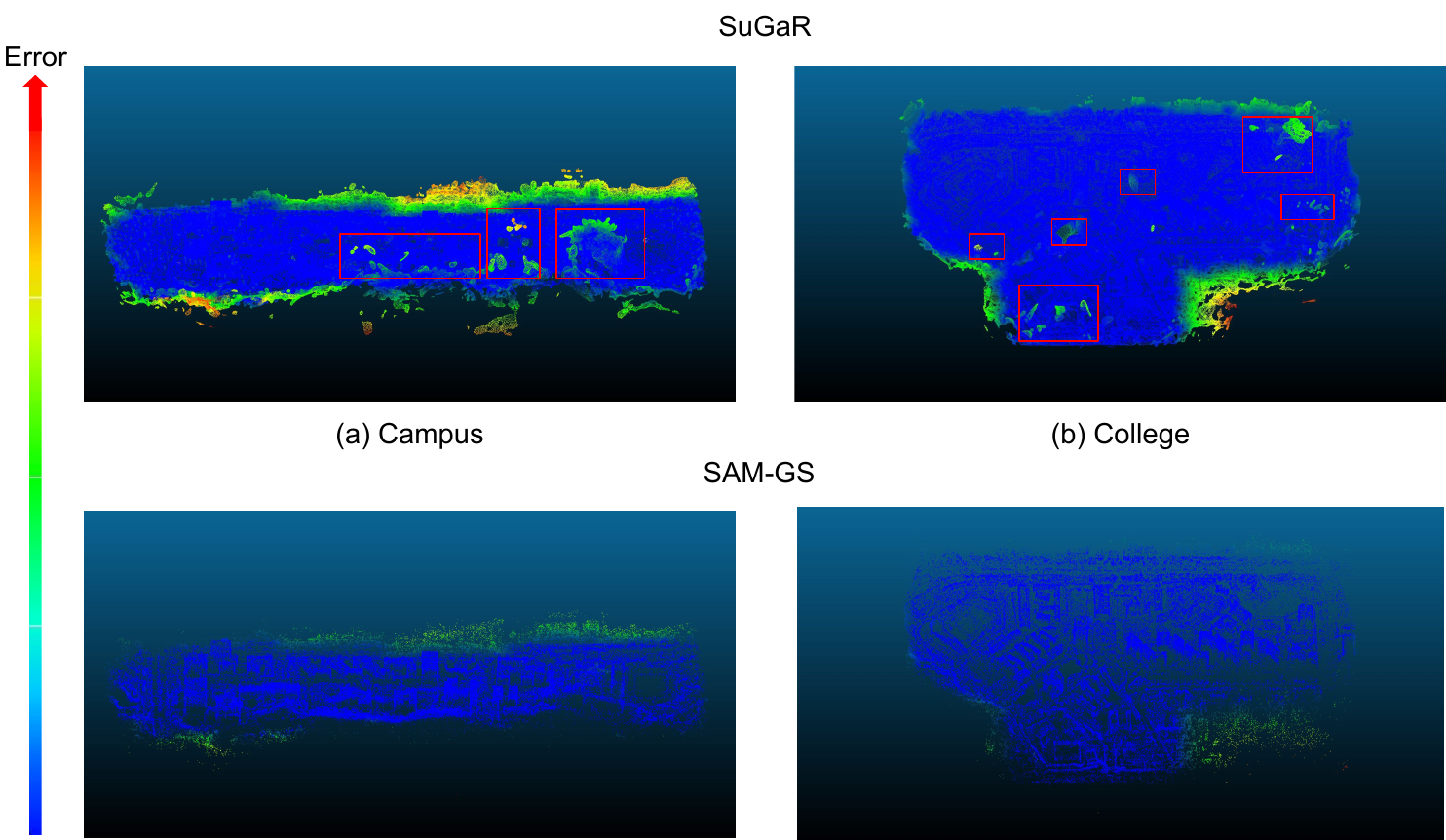}
    \caption{Explanation of Fantasy-Surface Problem: In the first row of this figure, we display the results of using SuGaR \cite{SuGaR} to reconstruct the Campus and College scenes from GauUsceneV2 \cite{GauU_V2}. Many surfaces incorrectly model the lighting conditions due to complex effects, such as how glass reflects sunlight at different angles and how clouds block sunlight. These imaginary surfaces, which do not represent the true surface, are regarded as fantasy surfaces. Our method, shown in the bottom rows, largely alleviates this problem, as evident in the figure. Another major source of geometric error occurs at the edges of unbounded scenes. However, this issue is common to all methods due to the sparsity of images at the edges and is not the focus of our current work.}
    \label{fig:fantasy_surface}
\end{figure*}

Despite the merits of 3D Gaussian Splatting (3DGS) mentioned above, it often suffers from unrealistic geometric reconstruction. Newly developed techniques primarily utilize both Signed Distance Functions (SDF) and other mesh-based reconstruction methods in conjunction with Gaussian representation to interactively train the system, ensuring that Gaussian splats adhere to the mesh surface. This approach has demonstrated improved quantitative performance in geometric reconstruction. However, this technique may generate random meshes when dealing with unbounded scenes, resulting in a larger chamfer distance compared to the ground truth LiDAR point cloud. Additionally, the works are based on the assumption that Gaussian splats will automatically adopt a disk-like shape. By aligning the normal of the disk, the extracted mesh is expected to be smoother. The work pushes this constraint further, as shown in the figure. These methods directly constrain the shape of Gaussians, degenerating 3D Gaussian Splats to 2D or using a shape constraint to encourage a disk shape Gaussian, facilitating mesh extraction by obtaining the normals of a 2D disk. However, curved surfaces cannot be effectively represented using disks. The strategy to generate curved surfaces typically results in the generation of a large number of 2D Gaussian Splats, leading to extremely high memory consumption. Most geometric methods naively encourage every Gaussian to have relatively high opacity to make Gaussians align with real objects' surfaces. Although this naive approach can work when there is stable lighting, complex lighting conditions can generate many small pieces of meshes, which are even worse than vanilla Gaussian Splatting as shown in the figure. In the current work, we regard this problem as a fantasy-surface problem. We further argue that simply encouraging a disk-shaped and large alpha value Gaussian Splat for every Gaussian could detrimentally impact overall reconstruction performance, as curved surfaces cannot be accurately represented by a disk-like shape and lead to a fantasy-surface problem.

To solve the problem mentioned above, we introduce a semantic-aware masked model as shown in the figure. We still take the same COLMAP input. But we control the shape of different Gaussians in a fine-grained style according to their semantic attributes. We first obtain the semantic attributes using GroundingSAM, and measure their geometric properties. We refer to all objects in the same category as a semantic group. The geometric property of a semantic group is termed geometric perplexity. The conclusion is that geometry complexity is related to the density of effective edges within a semantic group. To reach this conclusion, we provide a detailed frequency domain analysis in the method section.

A naive way to train is using the expected shape for each semantic group as a shape constraint to detail control the shape of each Gaussian during training and merge all semantic groups into one scene. However, this naive setting is not working due to the inconsistency of GroundingSAM. To be specific, in some images, some semantic parts are regarded as ground while for the consecutive image, the model will regard the same semantic part as something else as shown in the figure. We then provide a robust loss function served as a soft regularization to encourage each Gaussian Splat to form the shape we want. By first extending the geometric complexity idea to the whole model, we can then calculate the lower bounds of the number of 3DGS we need to fully construct the scene. Therefore, we provide a training strategy that iteratively decreases 3DGS during training, which reduces the training memory consumption.

We have observed that geometric reconstruction often conflicts with lighting effects. Specifically, when the reconstructed geometry aligns well with the point cloud, the image-based measurement matrix such as SSIM, LPIPS, and PSNR, tends to perform poorly. Simply enforcing all Gaussian has an alpha value larger than some specific opacity will lead to a fantasy-surface problem as we mentioned before. To address this, we introduce the hierarchical probability density sampling strategy. To be specific, low alpha value usually generates for complex lighting effect, by using or sampling strategy, the extracted geometry will not sample the low alpha value area. Therefore the extracted geometry is much better without harming the rendering result. Experiment results show that our method surpasses the current SOTA method such as SuGaR and 2D Gaussian Splats by a large margin.

\begin{figure*}
    \centering
    \includegraphics[width=\linewidth]{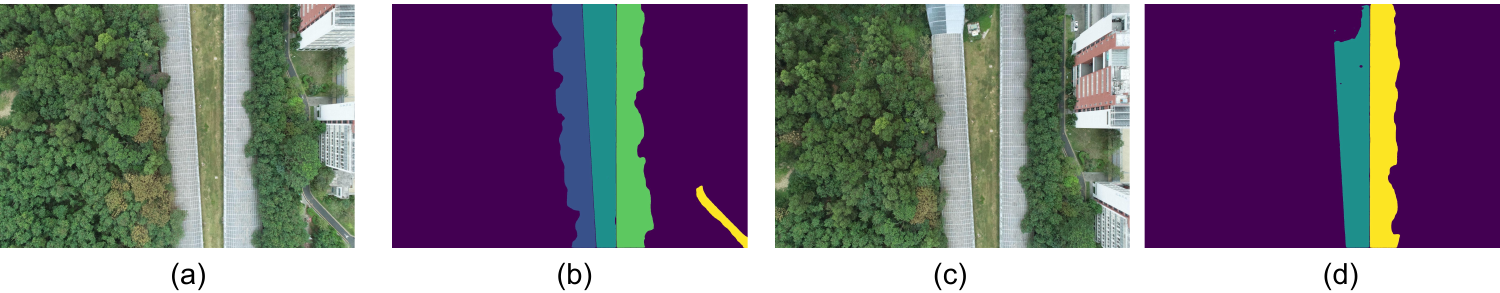}
    \caption{Explanation of Inconsistency problem. The semantic segmentation results are sometimes inconsistent with previous judgments. As shown in Figures (a) and (b), two tunnels are regarded as ground using GroundingSAM. However, in the images captured from a camera position immediately adjacent to them (Figures (c) and (d)), the left tunnel is not regarded as ground. This inconsistency between consecutive images is the primary cause of failure in naive reconstruction methods.}
    \label{fig:Inconsistancy}
\end{figure*}

\section{Related Work}
In this section, we introduce the Gaussian Splatting subsequent developments, particularly focusing on large-scale, geometric reconstruction and object level semantic aware 3DGS.

\subsection{Gaussian Splats on Large Scale Scene Reconstruction}
Large-scale scene reconstruction faces challenges such as high memory usage, variable lighting, and sparse data. Notable techniques include CityGaussian\cite{cityGS}, VastGaussian\cite{VastGaussian}, and HierarchicalGaussian\cite{hierachical3DGS}, which employ a divide-and-conquer strategy, though their methods differ. For instance, CityGaussian and VastGaussian segment based on camera visibility, while HierarchicalGaussian uses a grid division. Techniques like EfficientGaussian\cite{EfficientGS} focus on memory efficiency, introducing policies like gradient-sum thresholding and reducing low-impact Gaussians. Despite these advancements, geometric accuracy is still under-validated, lacking high-accuracy datasets. Recent datasets like GauUscene\cite{GauU_V2} and UrbanScene\cite{UrbanScene3D} offer more reliable data for testing. \cite{GauU_V2} highlights the mismatch between image-based metrics (SSIM, LPIPS, PSNR) and geometric metrics like Chamfer Distance, emphasizing the need for focused geometric assessments in our work.

\subsection{Gaussian Splats on Geometric Reconstruction}
Since the inception of 3D Gaussian Splatting \cite{vanilla_gaussian}, it has become a popular 3D representation method. However, aligning splats geometrically is challenging. Key advancements include SuGaR\cite{SuGaR} and ScaffoldGaussian\cite{scaffoldGaussian}, which improve surface alignment using specialized loss functions and neural networks, respectively. Techniques like 2DGS\cite{2DGS} simplify splats to 2D to ease mesh extraction. The novel SAGS approach enriches point clouds to enhance structural details. While these methods focus on shape regularization, our work integrates shape and semantic insights using GroundedSAM\cite{GroundingSAM} for enhanced geometric reconstruction.

\subsection{Object Level Semantic Aware Gaussian Splats}
Semantic-aware approaches like LangSplat\cite{LangSplat} and LeGaussian\cite{LeGaussian} leverage large vision models to understand object semantics through embeddings. While effective at the object level, our approach extends this understanding to large-scale scenes, utilizing semantic and geometric data to refine Gaussian Splats, showing marked improvements over purely geometric methods.

\section{SA-GS}
\begin{figure*}[htbp]
  \centering
    \includegraphics[width=\linewidth]{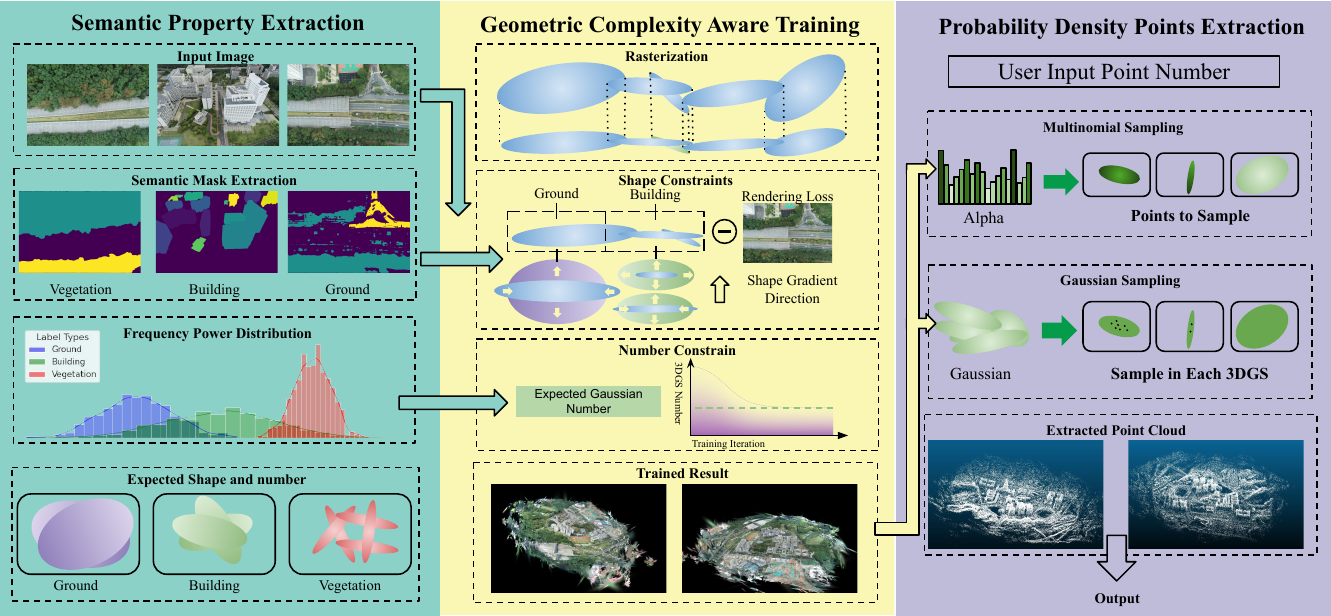}
   \caption{Method Overview: Our method pipeline consists of three main stages. Initially, we utilize the same input as vanilla Gaussian Splatting, but enhance it with semantic information extracted via Grounding SAM. Next, we assess the geometric complexity of each semantic group by calculating high-frequency power. Our geometric constraint is implemented through a soft regularization, facilitated by a semantic loss function. This guides the Gaussian shapes to match the expected shapes determined earlier. The rendering loss further refines the shape and attributes of the 3DGS, while the shape constraint, indicated by a negative sign, ensures alignment between rendered and real images. Controlling the shapes of different 3DGS is achieved by mapping their projected pixels onto the semantic map obtained earlier. Additionally, by reducing the number of low-opacity Gaussian splats to the expected count, we minimize GPU memory consumption during training. Finally, we offer a user-friendly point cloud extraction method via hierarchical probability density sampling. Initially, we create a multinomial distribution using the opacity values stored in each 3DGS. Then, based on user inputs and the multinomial distribution, we determine the number of points to sample from each Gaussian distribution. Detailed experimental results demonstrate significant improvements at each step, showcasing superior geometric reconstruction compared to current state-of-the-art methods.}
   \label{fig:method}
\end{figure*}
Our method has three overall stages as shown in Fig.\ref{fig:method}. At the first stage, we transfer the input images to different semantic masked outputs utilizing GroundingSAM\cite{GroundingSAM} and calculate the corresponding expected shape for each semantic group. At the second stage, we input the semantic map to start geometric complexity constraints training for Gaussian Splats. Notice that naively training different semantic groups separately will fail due to inconsistency, therefore we introduce a robust training and projection strategy at the training stage to solve this problem. In the third stage, we provide a hierarchical probability density extraction method for extracting the underlying geometry. In the following subsections, we will introduce each stage in detail.

\subsection{Geometric Property Extraction}
The major goal of the current stage is to obtain the expected shape of 3DGS for each semantic group. In the end, we find that the expected shape of the Gaussian is related to the number of edges contained in a semantic group. The expected shape extraction can be divided into two main sections: one for semantic mask extraction and one for power distribution mapping.
\subsubsection{Semantic Mask Extraction.}
We have $N$ input images $\mathcal{I} = \{I_1, I_2, \ldots, I_N\}$. We use GroundingSAM to get the masks of the images. The masked images are classified into different semantic groups. All masks are denoted as $\mathcal{M} \in \mathbb{R}^{H \times W \times N}$. We denote a pixel and position $(x, y)$ in image $i$ as $M_i(x, y)$. For every $M_i(x, y)$, it should be associated with a caption description or a default embedded value when GroundingSAM cannot find the correspondence. Caption embedding is denoted as $E(c)$, where $c$ is a text caption such as "vegetation", "buildings", "road", and so on. During implementation, the embedding $E(c)$ is set to integer numbers for simplicity. We denote $\forall M_i(x,y) = E(c_j)$ as a semantic group for caption $c_j$. We have an overall $C$ different captions. We then utilize the semantic group information as guidance to constrain the shape of the Gaussian.

\subsubsection{Power Distribution Mapping}
Gaussian distributions are often regarded as low-pass filters due to their tendency to favor low frequencies. In frequency domain analysis, the Gaussian Splatting model uses multiple low-frequency models to represent a general spectrum in 3D space. To represent high-frequency information, such as edges in 2D images or surfaces in 3D models, we use Gaussian splats with one relatively small axis to achieve high frequency along that axis. For low-frequency regions, we use larger Gaussians. Low-frequency areas cover a large space with fewer 3DGS, while high-frequency areas cover a small space but require more 3DGS. For semantic groups with low high-frequency information, we use disk-like shapes, and for groups with high-frequency signals, we use stick-like or dot-like Gaussians. High-frequency signals, often represented by edges or corners in images, are modeled using Canny edges. Mathematically, we define the expected ellipse with two aspect ratios.

\begin{align}
    a_1 = \frac{s_x}{s_z}, \quad    a_2 = \frac{s_y}{s_z}, 
\end{align}
where $s_x, s_y, s_z$ are the scale along x, y, and z axis. 
The detailed mathematical derivation for how to transfer from Gaussian shape to spectrum domain analysis and to edge extractor will be shown in the technical appendix Sec.\ref{Sec:Appendix}. One can examine it in detail if one wants.

During the real implementation, we extract the edge energy utilizing Canny Edge instead of simple high pass filter since Canny Edge can set the threshold for edge selection. Notice that Canny Edge is not the only edge extractor can be applied in our method, basiclly any edge extractor can finish the job. And we denote the edge count for image $i$ that in semantic group $j$ as $e_{ij}$

We define the overall perplexity of a certain semantic group as this:
\begin{align}
    \mathbf{P_j} = {\sum_i e_{ij}}, \quad i \in \mathcal{M}.
\end{align}
This perplexity is for whole semantic group within a scene. Therefore, we can obtain the expected Gaussian Splats number by multiply the edge number with a constant. This constant is usually determined by the overlapping ratio of the input images.  For unit perplexity, we need to divide by the number of pixel that in one semantic group. We denote it as $p_j$. 

Therefore, we have the following expectation for every Gaussian Splats in same semantic group: 
\begin{align}
    \frac{1}{\mathbf{p_i}} = k_1 a_1, \quad \frac{1}{P_i} = k_2 a_2, \quad k_1 > k_2.
\end{align}
\subsection{Geometric Complexity Aware Training}
To encourage Gaussian in different semantic group to align with the expected shape we have, we use the following loss function as geometric complexity loss function. 
\begin{align}
    \mathcal{L}_{gc} = \sigma(\frac{k_1}{ \mathbf{P_i} } -a_1) + \sigma(\frac{k_2}{ \mathbf{P_i} } -a_2).
\end{align}
As we mentioned before,we cannot directly train different semantic group separately due to the inconsistency issue. Therefore, we modify the CUDA kernel to extract the project mean of each Gaussian. We dynamically get the mask of Gaussian during training by attain the projected Gaussian location on 2D masks. Then we assign the expected shape to each Gaussian online. In this way, we circumvent the problem of inconsistency. In addition, since during rasterization process CUDA keep the projected mean of each Gaussians, our modified method still keep the same training time as memory consumption as vanilla Gaussian.

We adopt the same rendering loss as vanilla Nerf and use a hyper-parameter to tune the relationship between those loss as following: 
\begin{equation}
\lambda_{gc}\mathcal{L}_{gc} + \lambda_{dssim}\mathcal{L}_{dssim} + \lambda_{l1}\mathcal{L}_{l1}.
\end{equation}
Where $\mathcal{L}_{gc}$ is the geometry complexity loss. $\mathcal{L}_{dssim}$ and $\mathcal{L}_{l1}$ are the original rendering loss from vanilla 3DGS. We set $\lambda_{gc}, \lambda_{dssim}, \lambda_{l1}$ to 0.2, 0.2, and 0.6
We removes our Gaussian splats according to ranking, and it gradually decrease to the expected number linearly utilizing after first 6000 iteration. 

\subsection{Probability Density Point Extraction}
After obtained the trained Gaussian Spalts, we need to further extract the point cloud. Compare to previous method such as GaussianPro\cite{gaussian_pro}, GauUscene\cite{GauU_V2} using mean of each Gaussian as extracted point cloud we apply a hierachical probability distribution sampling strategy that best fit the geometric reconstruction requriment. 

We posit that the probability of a position containing a point is proportional to the product of Gaussian Density and opacity, given by:
\begin{align}
    \phi(x) = \sum_i{\alpha_i \exp\left(-\frac{1}{2}(\mathbf{x} - \boldsymbol{\mu}_i)^T \boldsymbol{\Sigma}_i^{-1} (\mathbf{x} - \boldsymbol{\mu}_i)\right)}.
    \label{Equ: distribution}
\end{align}

Notice the Equ.\ref{Equ: distribution} shows that the probability density function at position (x, y, z) is both proportional to Gaussian Probability density function and alpha value. Therefore we construct a two stage hierarchical sampling strategy as following. We first construct a multi-nominal distribution according to alpha value:
\begin{align}
    P(i) = \frac{\alpha_i}{\sum\alpha}.
\end{align}
By utilizing multi nominal distribution, we can first sample as many points as we want according to the alpha distribution. In reality, for index i with high alpha value, we might have multiple points fall inside, while for other index with relatively low alpha value, there might be no point fall inside. Then we sample the points within Gaussian. In this way we can sample as many points as we want. Experiment results shows superiority in geometric reconstrucion when applying the above point cloud extraction strategy. 

\section{Experiment}
\begin{figure*}[htbp]
  \centering
    \includegraphics[width=\linewidth]{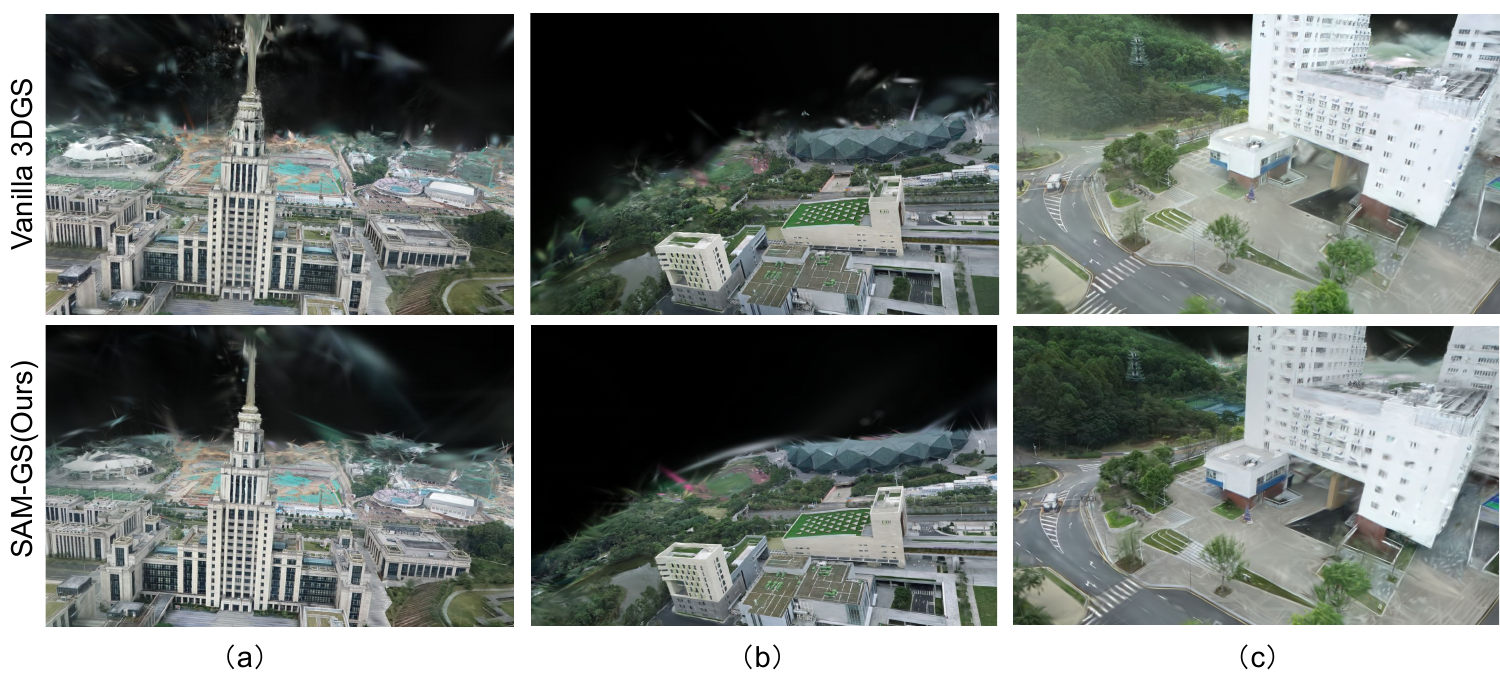}
   \caption{This the comparison between our method and vanilla Gaussian Splats. As one can see that from the figure shown above, our methods largely sharpening the edge of image. The tower shown in the figure (a) merges together and sharpened in our method, while in (b) figures, we eliminate the noise around the high building. While for the last group of pictures shows that our steadily alpha decreasing strategy is successful.}
   \label{fig:rendered_comaprison}
\end{figure*}

To validate our proposed method, we conducted intensive tests on both a public dataset, GauUScene V2, covering over 6.5 square kilometers, and our own "Technology Campus" dataset, spanning 1.8 square kilometers. Additionally, we used the UrbanScene3D Polytech scene (1.7 square kilometers) exclusively for image-based rendering validation due to its limited LiDAR data and lack of RGB information. Collectively, these datasets cover more than 10 square kilometers, demonstrating our method's effectiveness. 

\subsection{Implementation Detial and Metrics}
We compared our method against several leading techniques, including SuGaR, Vanilla Gaussian Splatting, InstantNGP, and NeRFacto, using the RTX 3090 for training. Our evaluation focused on geometric reconstruction, primarily using Chamfer distance, but also included image-based metrics such as PSNR, SSIM, and LPIPS for comprehensive assessment. The datasets and additional validation results will be made available in the supplementary materials.We train our model use RTX3090 and we set K1 to 3, k2 to 1. 
\subsection{Comparison}
Since real-world LiDAR data is usually smaller than the reconstructed region using NeRF-based and Gaussian-based methods, we crop our aligned reconstruction to the same size as the LiDAR data to maintain fairness. Due to page limitations, we only show the qualitative comparison between our method and the vanilla Gaussian. We display rendering results qualitatively in Fig.~\ref{fig:rendered_comaprison}. We then display our quantitative results on geometric alignment in Fig.\ref{fig:Geometric_Comparison}. Our geometric-based metric results are shown in Tab.\ref{tab:Chamful_distance}. As for image-based metrics, we sample 10\% of images from the whole dataset as the testing dataset, with the remaining data used for training. Since our training procedure strictly follows the GauScene V2 benchmark, all results on GauScene image-based metrics other than SA-GS are directly borrowed from their report. We find using one card to train the NeRF model is much faster than training it in parallel; therefore, we train all models using one RTX 3090. The detailed comparison is shown in Tab.\ref{tab:Image_based_matrix} and Tab.\ref{tab:Image_based_matrix_extend}

\begin{table*}[htbp]
\centering
\caption{Comparison on Chamful distance and variance. This table displays the results obtained when testing two NeRF-based methods and 3DGS (3D Gaussian Splatting). For training and evaluation, we used SuGaR and the official Gaussian Splats code. Meanwhile, the NeRF Studio implementation was utilized for Instant-NGP and NeRFacto to conduct training and evaluation.}
\label{tab:Chamful_distance}
\resizebox{1\linewidth}{!}{
\begin{tabular}{l|cc|cc|cc|cc|cc}
\toprule
Method 
& \multicolumn{2}{c|}   {Gaussian Splatting} 
& \multicolumn{2}{c|}   {SuGaR} 
& \multicolumn{2}{c|}   {SA-GS(Ours)} 
& \multicolumn{2}{c|}   {Instant NGP} 
& \multicolumn{2}{c}    {NeRFacto} 
\\
\begin{tabular}{c|c} Scene & Metrics \end{tabular}  
& Mean \(\downarrow\) & Var \(\downarrow\) 
& Mean \(\downarrow\) & Var \(\downarrow\) 
& Mean \(\downarrow\) & Var \(\downarrow\) 
& Mean \(\downarrow\) & Var \(\downarrow\) 
& Mean \(\downarrow\) & Var \(\downarrow\)  \\
\midrule
Campus  
        & 0.044 & 0.117
        & 0.081 & 0.163
        & \textbf{0.037} & \textbf{0.104}
        & 0.083 & 0.265
        & 0.054 & 0.169
        \\
Modern Building 
        & 0.040 & \textbf{0.101}
        & 0.066 & 0.140
        & 0.040 & 0.108
        & 0.065 & 0.186
        & \textbf{0.032} & 0.140
        \\
Village 
        & 0.039 & 0.087
        & 0.055 & 0.117
        & \textbf{0.035} & 0.081
        & 0.103 & 0.323
        & 0.036 & \textbf{0.077}
        \\
Residence 
        & 0.240 & 0.176
        & Nan   & Nan
        & \textbf{0.203} & \textbf{0.148}
        & 0.204 & 0.223
        & 0.263 & 0.226
        \\
Russian Building
        & 0.097 & 0.273
        & 0.186 & 0.390
        & 0.083 & 0.250
        & 0.056 & 0.167
        &\textbf{ 0.041} & \textbf{0.155}
        \\
College
        & 0.038 & 0.107
        & 0.058 & 0.139
        & 0.028 & 0.103
        & 0.087 & 0.270
        & \textbf{0.024} & \textbf{0.082}
        \\
Technology Campus
        & 0.073 & 0.229
        & 0.146 & 0.339
        & \textbf{0.056} & \textbf{0.204}
        & 0.057 & 0.166
        & 0.093 & 0.270
        \\
\hline
Avg. 
        & 0.076 & 0.155
        & 0.098 & 0.215
        & \textbf{0.068} & \textbf{0.143}
        & 0.094 & 0.229
        & 0.078 & 0.160
        \\
\bottomrule
\end{tabular}}
\end{table*}

\begin{table*}[htbp]
\centering

\caption{Image-based metrics result. The methods used are the same as mentioned above. We also provide additional training time for a detailed comparison. In the first six datasets, according to the GauUscene paper, they used four GPUs for training. Therefore, the last two datasets were trained on a single RTX 3090.}
\label{tab:Image_based_matrix}
\resizebox{1\linewidth}{!}{
\begin{tabular}{l|cccc|cccc}
\toprule
Method 
& \multicolumn{4}{c|}   {Instant NGP} 
& \multicolumn{4}{c}    {NeRFacto} 
\\
\begin{tabular}{c|c} Scene & Metrics \end{tabular}  
& PSNR \(\uparrow\) & SSIM \(\uparrow\) & LPIPS \(\downarrow\) & Time(GPU$\cdot$min) \(\downarrow\)
& PSNR \(\uparrow\) & SSIM \(\uparrow\) & LPIPS \(\downarrow\) & Time(GPU$\cdot$min) \(\downarrow\)
\\
\midrule
Campus  
        & 20.76 & 0.516 & 0.817 & 220 
        & 17.70 & 0.455 & 0.779 & 1692
        \\
Modern Building 
        & 20.25 & 0.522 & 0.816 & 392
        & 18.66 & 0.448 & 0.734 & 1704
        \\
Village 
        & 20.79 & 0.511 & 0.792 & 268 
        & 16.95 & 0.399 & 0.727 & 1788
        \\
Residence 
        & 18.64 & 0.453 & 0.856 & 348
        & 15.05 & 0.364 & 0.879 & 1780
        \\
Russian Building
        & 18.37 & 0.507 & 0.810 & 252
        & 16.61 & 0.405 & 0.682 & 1716
        \\
College
        & 19.64 & 0.551 & 0.820 & 276
        & 17.28 & 0.462 & 0.781 & 1732
        \\
Technology Campus
        & 19.58 & 0.510 & 0.720 & \textbf{38}
        & 17.78 & 0.463 & 0.805 & 685
        \\
polytech
        & 17.83 & 0.494 & 0.843 & \textbf{44}
        & 15.55 & 0.460 & 0.928 & 751
        \\
\hline
Avg. 
        & 19.73 & 0.508 & 0.809 & 230
        & 16.95 & 0.432 & 0.789 & 1481
        \\
\bottomrule
\end{tabular}}
\end{table*}

\begin{table*}[htbp]
\centering

\resizebox{1\linewidth}{!}{
\begin{tabular}{cccc|cccc|cccc}
\toprule
  \multicolumn{4}{c|}   {Vanilla Gaussian} 
& \multicolumn{4}{c|}   {SuGaR} 
& \multicolumn{4}{c}   {SA-GS(Ours)} 
\\
  PSNR \(\uparrow\) & SSIM \(\uparrow\) & LPIPS \(\downarrow\) & Time \(\downarrow\)
& PSNR \(\uparrow\) & SSIM \(\uparrow\) & LPIPS \(\downarrow\) & Time \(\downarrow\)
& PSNR \(\uparrow\) & SSIM \(\uparrow\) & LPIPS \(\downarrow\) & Time \(\downarrow\)
\\
\midrule
          24.76 & \textbf{0.735} & \textbf{0.343} & \textbf{58}  
        & 23.02 & 0.601 & 0.506 & 104
        &\textbf{25.16} & 0.730 & 0.363 & 68
        \\
          \textbf{25.49} & \textbf{0.762} & \textbf{0.273} & 64  
        & 22.51 & 0.572 & 0.497 & 108
        & 25.21 & 0.739 & 0.309 & \textbf{56}
        \\
          \textbf{26.14} & \textbf{0.805} & \textbf{0.237} & \textbf{62}  
        & 22.78 & 0.619 & 0.461 & 98  
        & 25.64 & 0.783 & 0.271 & 64
        \\
          \textbf{22.03} & \textbf{0.678} & \textbf{0.371} & \textbf{71}  
        & 20.97 & 0.533 & 0.607 & 119
        & 21.26 & 0.629 & 0.419 & 72
        \\
          \textbf{23.90} & \textbf{0.784} & \textbf{0.248} & \textbf{63} 
        & 21.58 & 0.618 & 0.450 & 103 
        & 23.86 & 0.771 & 0.273 & 68
        \\
          \textbf{24.21} & \textbf{0.749} & \textbf{0.326} & \textbf{68}
        & 22.02 & 0.588 & 0.514 & 123
        & 24.14 & 0.724 & 0.363 & \textbf{68}
        \\
          \textbf{23.94} & \textbf{0.786} & \textbf{0.223} & 69
        & 21.47 & 0.556 & 0.431 & 146
        & 23.05 & 0.741 & 0.327 & 73
        \\
          22.31 & 0.772 & 0.273 & 77
        & 20.98 & 0.569 & 0.487 & 173
        & \textbf{22.32} & \textbf{0.778} & \textbf{0.243} & 72
        \\
\hline
          \textbf{24.10} & \textbf{0.758} & \textbf{0.287} & \textbf{60}
        & 21.92 & 0.528 & 0.494 & 121.8
        & 23.82 & 0.736 & 0.321 & 66.9
        \\
\bottomrule
\end{tabular}}

\caption{This table shows the result we obtained using Gaussian Spalts based method}
\label{tab:Image_based_matrix_extend}
\end{table*}

\subsection{Ablation Study}
In this section, we clearly display how each module in our model is used and its effect on the final result. We have four groups of ablation studies. The first group involves shape constraint only, followed by shape constraint with alpha constraint, which means steadily decreasing the number of Gaussian Splats after the warm-up iteration. Notice that we did not apply hierarchical point sampling for these two Gaussian experiments; instead, we directly use the mean of each Gaussian to represent the extracted point cloud. The following experiments involve the extracted point cloud from shape-only constraint and the extracted point cloud from Gaussian Splats with alpha constraint. We use GauSceneV2 for testing. Since two of the representations are point clouds, we use Chamfer Distance as the measurement metric. The quantitative result is shown on \ref{fig:Geometric_Comparison}. The quantitative result is shown in Tab.\ref{tab:Ablation}

\begin{table*}[htbp]
\centering
\caption{Ablation results: As one can see, each part of our model is essential for achieving good geometric results. The result with alpha constraints and point cloud extraction yields the best performance in terms of CD Mean and CD variance.}
\label{tab:Ablation}
\resizebox{1\linewidth}{!}{
\begin{tabular}{l|cc|cc|cc|cc|cc}
\toprule
Method 
& \multicolumn{2}{c|}   {Gaussian Splatting} 
& \multicolumn{2}{c|}   {Shape Only} 
& \multicolumn{2}{c|}   {Alpha Constrain} 
& \multicolumn{2}{c|}   {Extracted Shpae Only} 
& \multicolumn{2}{c}    {Extracted Alpha Constrain} 
\\
\begin{tabular}{c|c} Scene & Metrics \end{tabular}  
& Mean \(\downarrow\) & Var \(\downarrow\) 
& Mean \(\downarrow\) & Var \(\downarrow\) 
& Mean \(\downarrow\) & Var \(\downarrow\) 
& Mean \(\downarrow\) & Var \(\downarrow\) 
& Mean \(\downarrow\) & Var \(\downarrow\)  \\
\midrule
Campus  
        & 0.044 & 0.117
        & 0.049 & 0.128
        & 0.042 & 0.114
        & 0.041 & 0.114
        & \textbf{0.037} & \textbf{0.104}
        \\
Modern Building 
        & 0.040 & 0.108
        & 0.035 & 0.081
        & 0.041 & 0.110
        & 0.041 & \textbf{0.108}
        &\textbf{0.040} &\textbf{0.108}
        \\
Village 
        & 0.039 & 0.087
        & 0.043 & 0.097
        & 0.038 & 0.088
        & 0.038 & 0.087
        & \textbf{0.035} & \textbf{0.081}
        \\
Residence 
        & 0.240 & 0.176
        & 0.265 & 0.188
        & 0.231 & 0.168
        & 0.209 & 0.156
        & \textbf{0.203} & \textbf{0.148}
        \\
Russian Building
        & 0.097 & 0.273
        & 0.105 & 0.296
        & 0.092 & 0.267
        & 0.093 & 0.278
        & \textbf{0.083} & \textbf{0.250}
        \\
College
        & 0.038 & 0.107
        & 0.047 & 0.122
        & 0.034 & 0.105
        & 0.032 & 0.108
        & \textbf{0.028} & \textbf{0.103}
        \\
\hline
Avg. 
        & 0.083 & 0.145
        & 0.091 & 0.167
        & 0.080 & 0.142
        & 0.076 & 0.142
        & \textbf{0.071} & \textbf{0.132}
        \\
\bottomrule
\end{tabular}}
\end{table*}

\section{Conclusion}
In our current work, we propose a semantic-aware geometric constraint algorithm that dynamically assigns expected shapes to Gaussian splats projected into different semantic groups. We present an algorithm capable of computing the geometric complexity of Gaussian splats based on spectrum analysis. Furthermore, we utilize geometric complexity measurement to determine the number of Gaussian splats. Subsequently, we introduce a hierarchical probability density sampling method that can extract as many points as desired by users while maintaining a dynamic alpha value to mitigate the fantasy surface problem. Additionally, we offer abundant experimental results. However, there are several drawbacks to our algorithm. Firstly, during training, we constrain the shape of all Gaussians that project onto the same pixel without explicitly ignoring Gaussians blocked by those with high opacity values before them. This may result in all Gaussians conforming to the shape of the semantic group that occupies the largest region in the scene. Secondly, our algorithm relies on key semantics provided by users, which may sometimes be absent. Thirdly, while the inconsistency between consecutive images can be addressed by our robust loss, the direct resolution of inconsistency in the 3D world itself has not been achieved.
\section{Acknowledgement}
This project is generously sponsored by \href{https://www.aukilabs.com/}{Auki Labs}. Auki Labs creates posemesh domains: virtual real estate for shared augmented reality, allowing you to manifest your knowledge and imagination in the minds of others. Thanks Zihao Fang for his great help on our visualization result.


\bibliographystyle{plain}
\bibliography{references}
\section{Appendix}
\subsection{Mathematical Derivation of Spectrum, Edge, and Gaussian Shape relationship}
\appendix
\label{Sec:Appendix}
Mathematically, we define the expected ellipse with two aspect ratios.

\begin{align}
    a_1 = \frac{s_x}{s_z},
    a_2 = \frac{s_y}{s_z}, 
\end{align}
where $s_x, s_y, s_z$ are the scale along x, y, and z axis. 
The detailed mathematical derivation for how to transfer from Gaussian shape to spectrum domain analysis and to edge extractor will be shown in the technical appendix. One can examine it in detail if one wants.
For each masked image $F(x,y)$ we transfer it to Fourier Domain as $F(u, v)$
A naive way to calculate how many Gaussian we need to use is to find the mean magnitude of frequency. When high frequency region is large, the mean magnitude of frequency will be large. To calculate magnitude of frequency we can first transfer $F(u, v)$ in to polar coordinates $f(\rho, \theta)$ and get the mean magnitude in Equ.\ref{Equ:Naive}, $f_\mu$ is magnitude of frequency. By calculating the weighted mean along the frequency orientation, we might find the geometric complexity.
\begin{align}
    f_\mu =  \int^{\frac{\pi}{2}}_{0}\int_{0}^{\infty}\rho\cdot |f(\rho, \theta)| d\rho d\theta 
    \label{Equ:Naive}
\end{align}
However, this naive magnitude statistics are not working well according to our experimental result. The major contribution of $f_\mu$ are actually low frequency information even though it has lower weight. We further observe that high frequency signal higher than certain threshold shown in image usually will lead to a dandified Gaussian Splats. Frequency higher than the threshold has the same effect on Gaussian Shape. This observation leads to a high-pass like statistic strategy as shown in Equ.\ref{Equ:High-pass}. Where $\delta$ function is an impulse function. 
\begin{align}
    f_\mu =  \int^{\frac{\pi}{2}}_{0}\int_{0}^{\infty}D(\rho)\cdot |f(\rho, \theta)| d\rho d\theta \quad D(\rho) = \delta(T - \rho)
    \label{Equ:High-pass}
\end{align}
Notice that the magnitude of spectrum $|f(\rho, \theta)|$ is positively related to the energy of spectrum $E(F(u, v) D(\rho))$. Notice that the energy itself is always the same. As shown in the Equ.\ref{Equ:energy}, Where the inverse Fourier transform of step function will be $\delta$ function with a phase shift. In here we simply use an impulse function to represent since we are calculating the real energy sum in the end. The left-hand-side of equation simply illustrate the energy of an image passing through a high pass filter.  
\begin{equation}
    E = \int^{\infty}_{-\infty} |{f(x, y) \ast \delta(x, y)|^2dxdy} \quad F^{-1}(D) = \delta(\rho)
    \label{Equ:energy}
\end{equation}
The effect of a high pass filter is actually an edge extraction kernel especially in image processing. During the real implementation, we extract the edge energy utilizing Canny Edge instead of simple high pass filter since Canny Edge can set the threshold for edge selection. And we denote the edge count for image $i$ that in semantic group $j$ as $e_{ij}$ 

We define the overall perplexity of a certain semantic group as this:
\begin{align}
    \mathbf{P_j} = {\sum_i e_{ij}}, \quad i \in \mathcal{M}
\end{align}
This perplexity is for whole semantic group within a scene. Therefore, we can obtain the expected Gaussian Splats number by multiply the edge number with a constant. This constant is usually determined by the overlapping ratio of the input images. That is when the overlapping ratio is large, the constant should be small. For unit perplexity, we need to divide by the number of pixel that in one semantic group. We denote it as $p_j$. 

Therefore, we have the following expectation for every Gaussian Splats in same semantic group: 
\begin{align}
    \frac{1}{\mathbf{p_i}} = k_1 a_1 \quad \frac{1}{P_i} = k_2 a_2 \quad k_1 > k_2
\end{align}
\end{document}